\pdfoutput=1

\documentclass[11pt]{article}

\usepackage{ACL2023}
\usepackage{booktabs}
\usepackage{times}
\usepackage{latexsym}

\usepackage[T1]{fontenc}

\usepackage[utf8]{inputenc}

\usepackage{microtype}

\usepackage{inconsolata}

\usepackage{graphicx}
\usepackage{hyperref}

\title{Empathy and Distress Detection using Ensembles of Transformer Models}

  \author{Tanmay Chavan\thanks{~ Equal contribution}~~, 
  Kshitij Deshpande$\footnotemark[1]$~ \and 
  Sheetal Sonawane$\footnotemark[1]$~ \\
  Pune Institute of Computer Technology, Pune \\
  \texttt{\{chavantanmay1402, kshitij.deshpande7\}@gmail.com, sssonawane@pict.edu}}

\begin{document}
\maketitle
\begin{abstract}
This paper presents our approach for the WASSA 2023 Empathy, Emotion and Personality Shared Task. Empathy and distress are human feelings that are implicitly expressed in natural discourses. Empathy and distress detection are crucial challenges in Natural Language Processing that can aid our understanding of conversations. The provided dataset consists of several long-text examples in the English language, with each example associated with a numeric score for empathy and distress. We experiment with several BERT-based models as a part of our approach. We also try various ensemble methods. Our final submission has a Pearson's r score of 0.346, placing us third in the empathy and distress detection subtask.
\end{abstract}


\section{Introduction}
Empathy and distress are important attributes in natural language processing which allow a better understanding of the nature of human interactions. However, they are difficult to quantify and as a result, are not as deeply researched as fields like hate and sentiment detection. Nevertheless, they are very beneficial in comprehending useful information. Thus, there is a huge scope for work to be done in this domain. Empathy is an emotion that enables us to grasp the emotional and mental state of others and thus is very crucial to conversations. It helps foster deeper social connections and promotes amicability. Hence, precisely identifying empathy is very beneficial. On the other hand, distress is an emotion that acts as a vital sign that suggests a possible threat or harm \cite{articledistress}. It thus helps identify discomfort and thereby makes efforts to allay any suffering that may have resulted. Thus, accurate distress detection helps promote well-being and peace in society.

The WASSA 2023 Empathy, Emotion and Personality Shared Task \citet{barriere2023wassa} was to perform Empathy Detection, Emotion Classification, and Personality Detection in Interactions. Our team, PICT-CLRL, participated under the Codalab username \textit{earendil} in  the shared task. 5 tracks, namely Empathy and Emotion Prediction in Conversations (CONV), Empathy Prediction (EMP), Emotion Classification (EMO), Personality Prediction (PER), and Interpersonal Reactivity Index Prediction (IRI) were offered. This paper demonstrates our work on Task 2 Empathy Prediction (EMP). Specifically, Task 2 was to predict the empathy and distress levels at an essay level. The essays supplied in the dataset are between 300 and 800 characters in length. These essays are written in response to news stories about people, groups, or other entities facing some sort of plights. Various attributes related to an individual demographic like age, gender, ethnicity, income, and education level are also provided. The average Pearson correlation for empathy and distress was considered the official metric for evaluation.

Recently, transformer-based models \citet{vaswani2017attention} have achieved great success in several NLP-related tasks. BERT \citet{devlin2019bert} has achieved State-of-the-art results on several benchmarks. Furthermore, several BERT-based models with additional pre-training have also produced excellent performance. The selection of pre-training data can aid domain-specific tasks as well. We experiment with several such models. In addition to this, we also try ensemble-based approaches. Ensembling can produce better results than individual models by combining model outputs in an effective manner. We evaluate these approaches and present the results and observations.

\begin{figure*}[t]
    \centering
    \includegraphics[width=\textwidth]{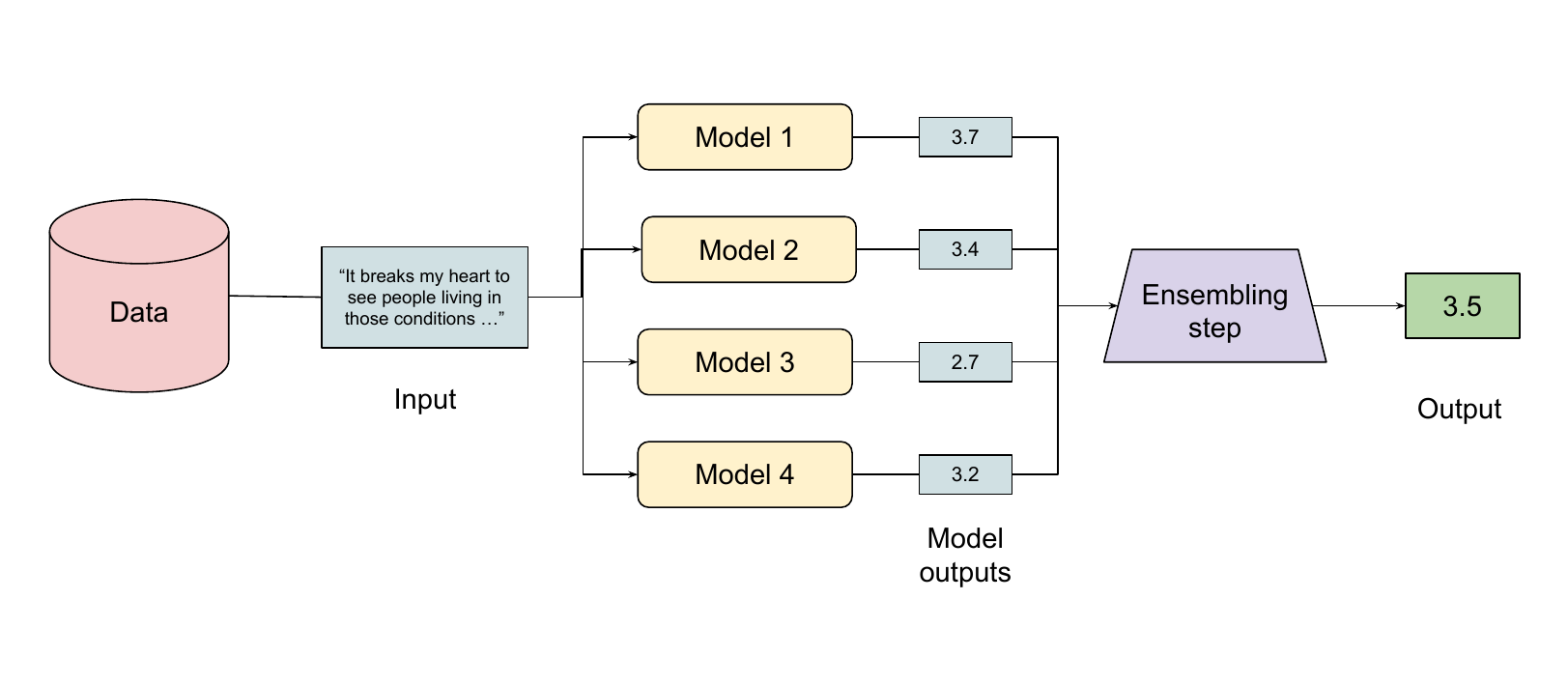}
    \caption{Our ensembling system. We use the same configuration for empathy and distress.}
    \label{fig:ensemble}
\end{figure*}

\section{Related Work}
\citet{litvak2016social} identifies that monitoring social and linguistic behavior through empathy monitoring hasn't gained much attention and that there is a huge scope for further research. To determine how social and linguistic behavior relates to the attribute of empathy, a Poisson regression model has been utilized. To better comprehend empathy, Davis' Interpersonal Reactivity Index (IRI), which takes into account 4 factors (namely fantasy, personal distress, empathetic concern, and perspective taking) has been used.\\
\citet{gibson2016deep} utilizes empathy in addiction counseling. The transcripts of the session conducted are used to train a model and predict empathy. Naturally, high empathy is desirable from a counselor toward the client. The model is trained in two parts, firstly a Recurrent Neural Network is trained on a group of certain behavioral acts and then this is used to train the final Deep Neural Network. This approach is shown to have produced better results than training the Deep neural network all at once.\\
\citet{hosseini2021distilling} identifies that it can be difficult to annotate data to identify empathy when working on a large scale. To integrate knowledge from the available resources and detect empathy from the natural language in several domains, this study uses multi-task training with knowledge distillation. Results on the TwittEmp dataset are shown to produce significantly better results using this approach.\\
\citet{saleem2012automatic} recognizes that psychological distress is seldom sufficiently identified. It offers ways to detect distress indicators and assess the severity of the distress. Text from online forum posts where individuals discuss their thoughts more freely is used. SVMs are used to identify distress indicators.\\
\citet{guda2021empathbert} utilizes user demographic to create an EMPATH-BERT framework for empathy detection. Internally it uses the BERT model, the framework is shown to surpass existing machine learning techniques. This paper allows us to understand the important role of demographic information in empathy detection.\\
\citet{barriere2022wassa} summarizes the previous edition of the shared task  and covers several approaches for the problem of empathy and distress prediction. 

\section{Data}
We use the \textit{Empathic Conversations} dataset presented by \citet{omitaomu2022empathic} for the task. The original data was split into 72\%, 18.9\%, and 9.1\% for train, dev, and test respectively. The data splits are summarized in Table \ref{tab:split}. Compared to the conversations dataset, the essay dataset is relatively smaller. The data present consists of 24 attributes. The training dataset has 792 examples. The validation set provided has 208 samples. Lastly, the testing dataset has 100 samples. The data consists of various individual demographic attributes like age, gender, and education along with essays that were composed in response to news stories about people experiencing hardships.

\begin{table}[]
\centering
\begin{tabular}{|c|c|lll}
\cline{1-2}
\textbf{Dataset} & \textbf{Number of examples} &  &  &  \\ \cline{1-2}
Training         & 792                       &  &  &  \\ \cline{1-2}
Validation       & 208                        &  &  &  \\ \cline{1-2}
Testing          & 100                        &  &  &  \\ \cline{1-2}
\end{tabular}
\caption{The dataset's training, validation, and test splits.}
\label{tab:split}
\end{table}

\begin{table*}[]
\centering
\begin{tabular}{|c|c|c|c|l}
\cline{1-4}
\textbf{Model}                                 & \textbf{\begin{tabular}[c]{@{}c@{}}Averaged \\ Pearson \\ Correlation\end{tabular}} & \textbf{\begin{tabular}[c]{@{}c@{}}Empathy \\ Pearson \\ Correlation\end{tabular}} & \textbf{\begin{tabular}[c]{@{}c@{}}Distress \\ Pearson \\ Correlation\end{tabular}} &  \\ \cline{1-4}
Twitter-RoBERTa-emotion          & 0.3189                                                                              & 0.3389                                                                    & 0.2991                                                                              &  \\
Twitter-RoBERTa-sentiment & 0.294                                                                               & 0.3128                                                                             & 0.2753                                                                              &  \\
Unsupervised SimCSE RoBERTa             & \textbf{0.35285}                                                                    & 0.3311                                                                             & \textbf{0.3746}                                                                     &  \\
RoBERTa-base                          & 0.29075                                                                             & \textbf{0.3444}                                                                             & 0.2371                                                                              &  \\ \cline{1-4}
\textbf{Ensembles}                             &                                                                                     &                                                                                    &                                                                                     &  \\
Mean                                  & \textbf{0.34619}                                                                    & 0.3585                                                                             & \textbf{0.3339}                                                                     &  \\
Linear Regression                     & 0.3285                                                                              & 0.3349                                                                             & 0.3221                                                                              &  \\
SVR                                   & 0.3221                                                                              & \textbf{0.3837}                                                                    & 0.2605                                                                              &  \\
XGBoost                               & 0.2898                                                                              & 0.3502                                                                             & 0.2294                                                                              &  \\ \cline{1-4}
\end{tabular}%
\caption{The results of our methods along with their scores. We use Pearson's \textit{r} as the scoring metric. }
\label{tab:results}
\end{table*}

\begin{table}[]
\centering
\begin{tabular}{|l|l|l|}
\hline
\textbf{Rank} & \textbf{Codalab ID} & \textbf{Score} \\ \hline
\textbf{1}    & ltm11               & 0.4178         \\
\textbf{2}    & Gruschka            & 0.3837         \\
\textbf{3}    & earendil            & 0.3462         \\
\textbf{4}    & zex                 & 0.3419         \\
\textbf{5}    & luxinxyz            & 0.3416         \\ \hline
\end{tabular}%
\caption{The top 5 teams at the EMP track. The score reported is averaged Pearson's correlation values of empathy and distress. Our team participated under the username \textit{earendil}.}
\label{tab:rankings}
\end{table}

\section{System Overview}

The shared task consists of a regression problem, where we have to predict a numerical value given an essay. BERT-based approaches are very successful at these types of problems. We try out several such LLMs. We also utilize various ensembling techniques. We briefly summarize our efforts in the following section.

\subsection{BERT-based Models}

We experiment with several pre-trained BERT-based models for the task. We evaluate their performances and select the models with the best performance on our dataset. 
RoBERTa\footnote{Model link: \href{https://huggingface.co/roberta-base}{https://huggingface.co/roberta-base}} \cite{liu2019roberta} is a BERT-based model with additional pre-training. It is pre-trained on 5 different English datasets, totaling a size of almost 160 GB. This is a vast improvement over BERT, which is pre-trained on two datasets totaling a size of about 16 GB. The authors claim that the additional pre-training results in improved performance, which is heavily supported by empirical evaluations. 

We also use Twitter-RoBERTa-emotion\footnote{Model link: \href{https://huggingface.co/cardiffnlp/twitter-roberta-base-emotion}{https://huggingface.co/cardiffnlp/twitter-roberta-base-emotion}} \cite{barbieri-etal-2020-tweeteval}. This model is a RoBERTa model which is pre-trained on roughly 58 million tweets. The model is also fine-tuned on the TweetEval benchmark datasets for emotion classification. 

Twitter-RoBERTa-sentiment\footnote{Model link: \\ \href{https://huggingface.co/cardiffnlp/twitter-roberta-base-sentiment-latest}{https://huggingface.co/cardiffnlp/twitter-roberta-base-sentiment-latest}} is another RoBERTa-based model used in our experiments. It pre-trains the roBERTa-base model on Twitter corpora composed of around 124 million tweets. The model is additionally fine-tuned on the TweetEval benchmark for sentiment analysis. The pre-training data for the model consists of tweets posted over a span of roughly 4 years, thus encompassing data spanning over a significant period of time.

The unsupervised SimCSE\footnote{Model link: \href{https://huggingface.co/princeton-nlp/unsup-simcse-roberta-base}{https://huggingface.co/princeton-nlp/unsup-simcse-roberta-base}} \cite{gao2022simcse} model uses sentence embeddings instead of token embeddings. It is trained in an unsupervised format and makes predictions on input sentences with a contrastive learning framework. Sentence-embedding-based models can potentially perform very well on long text-document classification tasks.

All of the above models are freely available on HuggingFace. We have tagged the model names with their links in the footnotes.

\subsection{Ensemble Methods}

Ensembling involves combining predictions by several individual models using various statistical and non-statistical-based approaches to enhance results. Ensembling can often result in better performances than individual models despite using the same data. We explore several ensembling approaches and elaborate on them.

The simplest ensembling method is to calculate average individual predictions and present them as the final output. Although this method might not utilize specific trends within the data, it generates stable predictions with low variance. 

We also use machine learning algorithms for ensembling outputs generated by the four models. We try linear regression and Support Vector Regression, as implemented in the sklearn module. We also use the XGBoost algorithm, an efficient variant of gradient boosting. Figure \ref{fig:ensemble} illustrates our ensemble system.

\section{Results}
\label{sec:results}

The results of our experiments are discussed in the following section. We report the results of the models and the ensembles in table \ref{tab:results}. The official scoring metric of the shared task for the EMP track is Pearson's \textit{r}. The final rank is determined by the average value of Pearson's correlations of empathy and distress.

We can see that the RoBERTa-base model has the highest score for predicting empathy. The unsupervised SimCSE performs best at predicting distress. Overall, the unsupervised SimCSE model has the best performance, with an average Pearson correlation of 0.352. The excellent performance of the SimCSE model is suggestive of the benefits of using sentence embeddings for long-text documents, in addition to better pre-training. It can also be observed that the Twitter-RoBERTa models do not exhibit significantly better performance than RoBERTA-base. This may be due to the fact that the Twitter pre-training data consisting of casually-written tweets, is significantly different in nature than the dataset responses which are properly and more mindfully composed.

Amongst the ensembling approaches, calculating the average value of individual model predictions generates the best result. It also outperforms other techniques at predicting distress. Support Vector Regression has the best results for empathy prediction. Although some BERT-based models perform better than others, supervised learning algorithms like linear regression and XGBoost fail to utilize this information and perform poorly. The poor performance can also be attributed to relatively less training data. 

Our team finished 3rd in the EMP track at the shared task. Our final submission has an average Pearson correlation score of 0.346. The top five participants along with their scores are reported in Table \ref{tab:rankings}.

\section{Conclusion}

The approach of Empathy Detection and Emotion Classification is proposed as part of the WASSA 2023 Empathy, Emotion and Personality Shared Task. Various methods are explored for the task. We have implemented several BERT-based models and evaluated them. We observed that the unsupervised SimCSE model has the best performance among the models we evaluated. It can also be seen that averaging the results of individual models generates the best results among the ensembling methods. Our final submission, with a Pearson's \textit{r} score of 0.346, is the third-best score in the EMP (Empathy Prediction) track at the shared task. Several improvements and future lines of work can be identified. Additional models pre-trained on relevant data can potentially boost methods. Furthermore, other ensembling techniques can also be explored and evaluated for better results.

\section*{Acknowledgment}

We thank the Pune Institute of Computer Technology's  Computational Linguistics Research Lab for providing the necessary help and guidance for the work. We are grateful for their support.

\section*{Limitations}
Although our models perform efficiently on the provided dataset, they might not be feasible in real-world scenarios due to the amount of computation they require. The development of more efficient models will vastly improve the deployability of such systems.



\bibliography{anthology,custom}
\bibliographystyle{acl_natbib}

\appendix

\end{document}